\def\year{2021}\relax
\documentclass[letterpaper]{article} 
\usepackage{aaai21}  
\usepackage{times}  
\usepackage{helvet} 
\usepackage{courier}  
\usepackage[hyphens]{url}  
\usepackage{graphicx} 
\usepackage{natbib}  
\usepackage{caption} 
\usepackage{amsfonts}
\usepackage{amsmath}
\usepackage{multirow}
\usepackage{subfigure}
\usepackage{pinyin}
\usepackage{tikz}
\usepackage{pgfplots}
\usetikzlibrary{plotmarks}
\usetikzlibrary{arrows}
\usetikzlibrary{decorations}
\usetikzlibrary{decorations.pathreplacing}
\usetikzlibrary{backgrounds}
\usetikzlibrary{fit}
\usetikzlibrary{positioning}
\usetikzlibrary{calc}
\usetikzlibrary{patterns}
\usetikzlibrary{shadows}
\usetikzlibrary[shapes.multipart]
\usetikzlibrary{pgfplots.groupplots}
\definecolor{ugreen}{cmyk}{1,0,1,0.498}
\definecolor{lyyblue}{cmyk}{0.8278,0.3333,0,0.2941}
\definecolor{lyygreen}{cmyk}{0.6813,0,0.725,0.3725}
\definecolor{lyyred}{cmyk}{0,0.8855,0.8767,0.1098}
\definecolor{dblue}{cmyk}{1,0.5487,0,0.5569}
\newcommand{\tab}[1]{Table \ref{#1}}%
\newcommand{\fig}[1]{Fig. \ref{#1}}%
\newcommand{\eqn}[1]{Eq. \ref{#1}}%
\newcommand{\my}{\textsc{Can}}%
\newcommand{\SAN}{\textsc{San}}%
\newcommand{\aan}{\textsc{Aan}}%
\title{An Efficient Transformer Decoder with Compressed Sub-layers}

\author{
Yanyang Li\textsuperscript{\rm 1}\thanks{Authors contributed equally.}, 
Ye Lin\textsuperscript{\rm 1}$^{*}$, 
Tong Xiao\textsuperscript{\rm 1,2}\thanks{Corresponding author.}, 
Jingbo Zhu\textsuperscript{\rm 1,2}\\
}
\affiliations{

    \textsuperscript{\rm 1}NLP Lab, School of Computer Science and Engineering,
    Northeastern University, Shenyang, China\\
    \textsuperscript{\rm 2}NiuTrans Research, Shenyang, China\\

    \{blamedrlee, linye2015\}@outlook.com,
    \{xiaotong,zhujingbo\}@mail.neu.edu.cn

}

\begin{document}

\maketitle

\begin{abstract}
    The large attention-based encoder-decoder network (Transformer) has become prevailing recently due to its effectiveness. But the high computation complexity of its decoder raises the inefficiency issue. By examining the mathematic formulation of the decoder, we show that under some mild conditions, the architecture could be simplified by compressing its sub-layers, the basic building block of Transformer, and achieves a higher parallelism. We thereby propose \emph{Compressed Attention Network}, whose decoder layer consists of only one sub-layer instead of three. Extensive experiments on 14 WMT machine translation tasks show that our model is 1.42$\times$ faster with performance on par with a strong baseline. This strong baseline is already 2$\times$ faster than the widely used standard baseline without loss in performance.
\end{abstract}

\section{Introduction}

Transformer is an attention-based encoder-decoder model \cite{DBLP:conf/nips/VaswaniSPUJGKP17}. It has shown promising results in machine translation tasks recently \cite{DBLP:conf/acl/WangLXZLWC19,DBLP:conf/emnlp/LiWLJDXWZ20,DBLP:conf/wmt/ZhangWCWSZRZZWM20,DBLP:conf/aaai/LiWXLZ20}. Nonetheless, Transformer suffers from the inefficiency issue at inference. This problem is attributed to the Transformer decoder for two reasons: 1) the decoder is deep \cite{DBLP:journals/corr/abs-2006-10369}. It consists of multiple layers and each layer contains three sub-layers, including two attentions and a feed-forward network; 2) the attention has a high (quadratic time) complexity \cite{DBLP:conf/acl/XiongZS18}, as it needs to compute the correlation between any two input words.

Previous work has focused on improving the complexity of the attention in the decoder to accelerate the inference. For example, \aan{} uses the averaging operation to avoid computing the correlation between input words \cite{DBLP:conf/acl/XiongZS18}. \SAN{} share the attention results among layers \cite{DBLP:conf/ijcai/XiaoLZ0L19}. On the other hand, we learn that vanilla attention runs faster in training than in inference thanks to its parallelism. This offers a new direction: a higher degree of parallelism could speed up the inference. The most representative work of this type is the non-autoregressive approach \cite{DBLP:conf/iclr/Gu0XLS18}. Its decoder predicts all words in parallel, but fails to model the word dependencies. Despite of their successes, all these systems still have a deep decoder.

In this work, we propose to parallelize the sub-layers to obtain a shallow autoregressive decoder. This way does not suffer from the poor result of directly reducing depths and avoids the limitation of non-autoregressive approaches. We prove that the two attention sub-layers in a decoder layer could be parallelized if we assume their inputs are close to each other. This assumption holds and thereby we compress these two attentions into one. Furthermore, we show that the remaining feed-forward network could also be merged into the attention due to their linearity. To the end, we propose \emph{Compressed Attention Network} (\my{} for short). The decoder layer of \my{} possesses a single attention sub-layer that does the previous three sub-layers' jobs in parallel. As another ``bonus'', \my{} is simple and easy to be implemented.

In addition, \citet{DBLP:journals/corr/abs-2006-10369} empirically discover that existing systems are not well balancing the encoder and decoder depths. Based on their work, we build a system with a deep encoder and a shallow decoder, which is 2$\times$ faster than the widely used standard baseline without loss in performance. It requires neither the architecture modification nor adding extra parameters. This system serves as a stronger baseline for a more convincing comparison.

We evaluate \my{} and the stronger baseline in 14 machine translation tasks, including WMT14 English$\leftrightarrow$\{German, French\} (En$\leftrightarrow$\{De, Fr\}) and WMT17 English$\leftrightarrow$\{German, Finnish, Latvian, Russian, Czech\} (En$\leftrightarrow$\{De, Fi, Lv, Ru, Cs\}). The experiments show that \my{} is up to 2.82$\times$ faster than the standard baseline with almost no loss in performance. Even comparing to our stronger baseline, \my{} still has a 1.42$\times$ speed-up, while other acceleration techniques such as \SAN{} and \aan{} are 1.12$\sim$1.16$\times$ in the same case.

\begin{figure*}[t!]
    \centering
    \hspace*{\fill}
    \subfigure[Transformer]
    {
        \centering
        \begin{tikzpicture}[scale=1.5]
            \tikzstyle{reprnode} = [rectangle,draw,rounded corners=2pt,minimum height=0.2cm,minimum width=0.6cm,inner sep=3pt]
            \tikzstyle{wordnode} = [minimum width=0.6cm,font=\small,align=center,inner sep=1pt]

            \begin{scope}[local bounding box=ENCODER]
                \node[reprnode,fill=ugreen!50!white,anchor=west] (enc1) at (0,0) {};
                \node[reprnode,fill=ugreen!50!white,anchor=west] (enc2) at ([xshift=3pt]enc1.east) {};
                \node[reprnode,fill=ugreen!50!white,anchor=west] (enc3) at ([xshift=3pt]enc2.east) {};
                \node[reprnode,fill=ugreen!50!white,anchor=west] (enc4) at ([xshift=3pt]enc3.east) {};

                \node[wordnode,anchor=north] (sw1) at ([yshift=-1.1cm]enc1.south) {w\o3};
                \node[wordnode,anchor=base west] (sw2) at ([xshift=3pt]sw1.base east) {h\en3};
                \node[wordnode,anchor=base west] (sw3) at ([xshift=3pt]sw2.base east) {h\ao3};
                \node[wordnode,anchor=base west] (sw4) at ([xshift=3pt]sw3.base east) {.};

                \node[reprnode,minimum height=0.6cm,opacity=0,anchor=north] (es1) at ([yshift=-0.3cm]enc1.south) {};
                \node[reprnode,minimum height=0.6cm,opacity=0,anchor=north] (es2) at ([yshift=-0.3cm]enc2.south) {};
                \node[reprnode,minimum height=0.6cm,opacity=0,anchor=north] (es3) at ([yshift=-0.3cm]enc3.south) {};
                \node[reprnode,minimum height=0.6cm,opacity=0,anchor=north] (es4) at ([yshift=-0.3cm]enc4.south) {};
                \node[rectangle,draw,rounded corners=2pt,fill=gray!10!white,inner sep=0pt,fit=(es1) (es4)] (encoder) {};
                \node[font=\small] () at (encoder) {Encoder};

                \foreach \i in {1,...,4}
                {
                    \draw[-latex,thick,black] (es\i.north) to (enc\i.south);
                    \draw[-latex,thick,black] ([yshift=-0.3cm]es\i.south) to (es\i.south);
                }
            \end{scope}

            \begin{scope}[local bounding box=DECODER]
                \node[reprnode,fill=dblue!50!white,anchor=west] (cross1) at ([xshift=0.6cm]enc4.east) {};
                \node[reprnode,fill=dblue!50!white,anchor=west] (cross2) at ([xshift=3pt]cross1.east) {};
                \node[reprnode,fill=dblue!50!white,anchor=west] (cross3) at ([xshift=3pt]cross2.east) {};
                \node[reprnode,fill=dblue!50!white,anchor=west] (cross4) at ([xshift=3pt]cross3.east) {};

                \node[reprnode,fill=dblue!50!white,anchor=north] (self1) at ([yshift=-0.6cm]cross1.south) {};
                \node[reprnode,fill=dblue!50!white,anchor=north] (self2) at ([yshift=-0.6cm]cross2.south) {};
                \node[reprnode,fill=dblue!50!white,anchor=north] (self3) at ([yshift=-0.6cm]cross3.south) {};
                \node[reprnode,fill=dblue!50!white,anchor=north] (self4) at ([yshift=-0.6cm]cross4.south) {};

                \node[reprnode,fill=dblue!50!white,anchor=south] (ffn1) at ([yshift=0.6cm]cross1.north) {};
                \node[reprnode,fill=dblue!50!white,anchor=south] (ffn2) at ([yshift=0.6cm]cross2.north) {};
                \node[reprnode,fill=dblue!50!white,anchor=south] (ffn3) at ([yshift=0.6cm]cross3.north) {};
                \node[reprnode,fill=dblue!50!white,anchor=south] (ffn4) at ([yshift=0.6cm]cross4.north) {};

                \node[reprnode,fill=dblue!50!white,anchor=south] (output1) at ([yshift=0.6cm]ffn1.north) {};
                \node[reprnode,fill=dblue!50!white,anchor=south] (output2) at ([yshift=0.6cm]ffn2.north) {};
                \node[reprnode,fill=dblue!50!white,anchor=south] (output3) at ([yshift=0.6cm]ffn3.north) {};
                \node[reprnode,fill=dblue!50!white,anchor=south] (output4) at ([yshift=0.6cm]ffn4.north) {};

                \node[wordnode,anchor=north] (tw1) at ([yshift=-0.3cm]self1.south) {I};
                \node[wordnode,anchor=base west] (tw2) at ([xshift=3pt]tw1.base east) {am};
                \node[wordnode,anchor=base west] (tw3) at ([xshift=3pt]tw2.base east) {fine};
                \node[wordnode,anchor=base west] (tw4) at ([xshift=3pt]tw3.base east) {.};

                \node[wordnode,anchor=south] (ow1) at ([yshift=0.3cm]output1.north) {am};
                \node[wordnode,anchor=base west] (ow2) at ([xshift=3pt]ow1.base east) {fine};
                \node[wordnode,anchor=base west] (ow3) at ([xshift=3pt]ow2.base east) {.};
                \node[wordnode,anchor=base west] (ow4) at ([xshift=3pt]ow3.base east) {$\langle$eos$\rangle$};

                \foreach \i in {1,...,4}
                {
                    \draw[-latex,thick,black] ([yshift=-0.3cm]self\i.south) to (self\i.south);
                    \draw[-latex,thick,black] (output\i.north) to ([yshift=0.3cm]output\i.north);
                }

                \draw[-latex,dblue,thick] (self1.north) .. controls +(north:0.2cm) and +(south:0.4cm) .. (cross4.south);
                \draw[-latex,dblue,thick] (self2.north) .. controls +(north:0.1cm) and +(south:0.4cm) .. (cross4.south);
                \draw[-latex,dblue,thick] (self3.north) .. controls +(north:0.1cm) and +(south:0.4cm) .. (cross4.south);
                \draw[-latex,dblue,thick] (self4.north) to (cross4.south);

                \node[wordnode,text=dblue,anchor=south west] () at ([yshift=0.3cm]self1.north west) {Self-Attention};

                \draw[-latex,ugreen,thick] (enc1.north) .. controls +(north:0.4cm) and +(south:0.5cm) .. (ffn4.south);
                \draw[-latex,ugreen,thick] (enc2.north) .. controls +(north:0.3cm) and +(south:0.5cm) .. (ffn4.south);
                \draw[-latex,ugreen,thick] (enc3.north) .. controls +(north:0.2cm) and +(south:0.5cm) .. (ffn4.south);
                \draw[-latex,ugreen,thick] (enc4.north) .. controls +(north:0.1cm) and +(south:0.5cm) .. (ffn4.south);
                \draw[-latex,ugreen,thick] (cross4.north) to (ffn4.south);
                
                \node[wordnode,text=ugreen,anchor=south west] () at ([yshift=0.3cm]enc2.north west) {Cross-Attention};

                \draw[-latex,blue,thick] (ffn4) to node [midway,left,text=blue,font=\small] {FFN} (output4);
            \end{scope}

            \begin{pgfonlayer}{background}
                \node[rectangle,draw,fill=gray!10!white,rounded corners=2pt,inner sep=2pt,fit=(self1) (output4),label={[font=\small]right:$\times N$}] (declayer) {};
            \end{pgfonlayer}

            \node[reprnode,fill=dblue!50!white,anchor=south west,label={[font=\small]right:Target Token}] (label1) at ([yshift=1cm]enc1.north west) {};
            \node[reprnode,fill=ugreen!50!white,anchor=south west,label={[font=\small]right:Source Token}] (label2) at ([yshift=5pt]label1.north west) {};
            \node[reprnode,fill=gray!10!white,anchor=south west,label={[font=\small]right:Decoder}] (label3) at ([yshift=5pt]label2.north west) {};
            \draw[-latex,draw,thick,densely dotted] ([xshift=1cm]label3.east) .. controls +(east:1cm) and +(west:1cm) .. ([yshift=0.3cm]declayer.west);
        \end{tikzpicture}
        \label{fig:transformer}
    }
    \hfill
    \subfigure[\my{}]
    {
        \centering
        \begin{tikzpicture}[scale=1.5]
            \tikzstyle{reprnode} = [rectangle,draw,rounded corners=2pt,minimum height=0.2cm,minimum width=0.6cm,inner sep=3pt]
            \tikzstyle{wordnode} = [minimum width=0.6cm,font=\small,align=center,inner sep=1pt]

            \begin{scope}[local bounding box=ENCODER]
                \node[reprnode,fill=ugreen!50!white,anchor=west] (enc1) at (0,0) {};
                \node[reprnode,fill=ugreen!50!white,anchor=west] (enc2) at ([xshift=3pt]enc1.east) {};
                \node[reprnode,fill=ugreen!50!white,anchor=west] (enc3) at ([xshift=3pt]enc2.east) {};
                \node[reprnode,fill=ugreen!50!white,anchor=west] (enc4) at ([xshift=3pt]enc3.east) {};

                \node[wordnode,anchor=north] (sw1) at ([yshift=-1.1cm]enc1.south) {w\o3};
                \node[wordnode,anchor=base west] (sw2) at ([xshift=3pt]sw1.base east) {h\en3};
                \node[wordnode,anchor=base west] (sw3) at ([xshift=3pt]sw2.base east) {h\ao3};
                \node[wordnode,anchor=base west] (sw4) at ([xshift=3pt]sw3.base east) {.};

                \node[reprnode,minimum height=0.6cm,opacity=0,anchor=north] (es1) at ([yshift=-0.3cm]enc1.south) {};
                \node[reprnode,minimum height=0.6cm,opacity=0,anchor=north] (es2) at ([yshift=-0.3cm]enc2.south) {};
                \node[reprnode,minimum height=0.6cm,opacity=0,anchor=north] (es3) at ([yshift=-0.3cm]enc3.south) {};
                \node[reprnode,minimum height=0.6cm,opacity=0,anchor=north] (es4) at ([yshift=-0.3cm]enc4.south) {};
                \node[rectangle,draw,rounded corners=2pt,fill=gray!10!white,inner sep=0pt,fit=(es1) (es4)] (encoder) {};
                \node[font=\small] () at (encoder) {Encoder};

                \foreach \i in {1,...,4}
                {
                    \draw[-latex,thick,black] (es\i.north) to (enc\i.south);
                    \draw[-latex,thick,black] ([yshift=-0.3cm]es\i.south) to (es\i.south);
                }
            \end{scope}

            \begin{scope}[local bounding box=DECODER]
                \node[reprnode,fill=dblue!50!white,anchor=west] (cross1) at ([xshift=0.6cm]enc4.east) {};
                \node[reprnode,fill=dblue!50!white,anchor=west] (cross2) at ([xshift=3pt]cross1.east) {};
                \node[reprnode,fill=dblue!50!white,anchor=west] (cross3) at ([xshift=3pt]cross2.east) {};
                \node[reprnode,fill=dblue!50!white,anchor=west] (cross4) at ([xshift=3pt]cross3.east) {};

                \node[reprnode,fill=dblue!50!white,anchor=south] (output1) at ([yshift=0.6cm]cross1.north) {};
                \node[reprnode,fill=dblue!50!white,anchor=south] (output2) at ([yshift=0.6cm]cross2.north) {};
                \node[reprnode,fill=dblue!50!white,anchor=south] (output3) at ([yshift=0.6cm]cross3.north) {};
                \node[reprnode,fill=dblue!50!white,anchor=south] (output4) at ([yshift=0.6cm]cross4.north) {};

                \node[wordnode,anchor=north] (tw1) at ([yshift=-0.3cm]cross1.south) {I};
                \node[wordnode,anchor=base west] (tw2) at ([xshift=3pt]tw1.base east) {am};
                \node[wordnode,anchor=base west] (tw3) at ([xshift=3pt]tw2.base east) {fine};
                \node[wordnode,anchor=base west] (tw4) at ([xshift=3pt]tw3.base east) {.};

                \node[wordnode,anchor=south] (ow1) at ([yshift=0.3cm]output1.north) {am};
                \node[wordnode,anchor=base west] (ow2) at ([xshift=3pt]ow1.base east) {fine};
                \node[wordnode,anchor=base west] (ow3) at ([xshift=3pt]ow2.base east) {.};
                \node[wordnode,anchor=base west] (ow4) at ([xshift=3pt]ow3.base east) {$\langle$eos$\rangle$};

                \foreach \i in {1,...,4}
                {
                    \draw[-latex,thick,black] ([yshift=-0.3cm]cross\i.south) to (cross\i.south);
                    \draw[-latex,thick,black] (output\i.north) to ([yshift=0.3cm]output\i.north);
                }

                \draw[-latex,red,thick] (cross1.north) .. controls +(north:0.2cm) and +(south:0.4cm) .. (output4.south);
                \draw[-latex,red,thick] (cross2.north) .. controls +(north:0.1cm) and +(south:0.4cm) .. (output4.south);
                \draw[-latex,red,thick] (cross3.north) .. controls +(north:0.1cm) and +(south:0.4cm) .. (output4.south);
                \draw[-latex,red,thick] (cross4.north) to (output4.south);

                \draw[-latex,red,thick] (enc1.north) .. controls +(north:0.5cm) and +(south:0.4cm) .. (output4.south);
                \draw[-latex,red,thick] (enc2.north) .. controls +(north:0.4cm) and +(south:0.4cm) .. (output4.south);
                \draw[-latex,red,thick] (enc3.north) .. controls +(north:0.3cm) and +(south:0.4cm) .. (output4.south);
                \draw[-latex,red,thick] (enc4.north) .. controls +(north:0.2cm) and +(south:0.4cm) .. (output4.south);
                
                \node[wordnode,text=red,anchor=south west] () at ([yshift=0.35cm]enc1.north west) {Compressed-Attention};
            \end{scope}

            \begin{pgfonlayer}{background}
                \node[rectangle,draw,fill=gray!10!white,rounded corners=2pt,inner sep=2pt,fit=(cross1) (output4),label={[font=\small]right:$\times N$}] (declayer) {};
            \end{pgfonlayer}

            \node[reprnode,fill=dblue!50!white,anchor=south west,label={[font=\small]right:Target Token}] (label1) at ([yshift=1cm]enc1.north west) {};
            \node[reprnode,fill=ugreen!50!white,anchor=south west,label={[font=\small]right:Source Token}] (label2) at ([yshift=5pt]label1.north west) {};
            \node[reprnode,fill=gray!10!white,anchor=south west,label={[font=\small]right:Decoder}] (label3) at ([yshift=5pt]label2.north west) {};
            \draw[-latex,draw,thick,densely dotted] ([xshift=1cm]label3.east) .. controls +(east:1cm) and +(west:1cm) .. ([yshift=0.3cm]declayer.west);
        \end{tikzpicture}
        \label{fig:man}
    }
    \hspace*{\fill}
    \caption{Transformer vs. \my{} (Chinese pinyin$\rightarrow$English: \emph{``w\o3 h\en3 h\ao3 .''} $\to$ \emph{``I am fine .''}).}
    \label{fig:comparison}
\end{figure*}
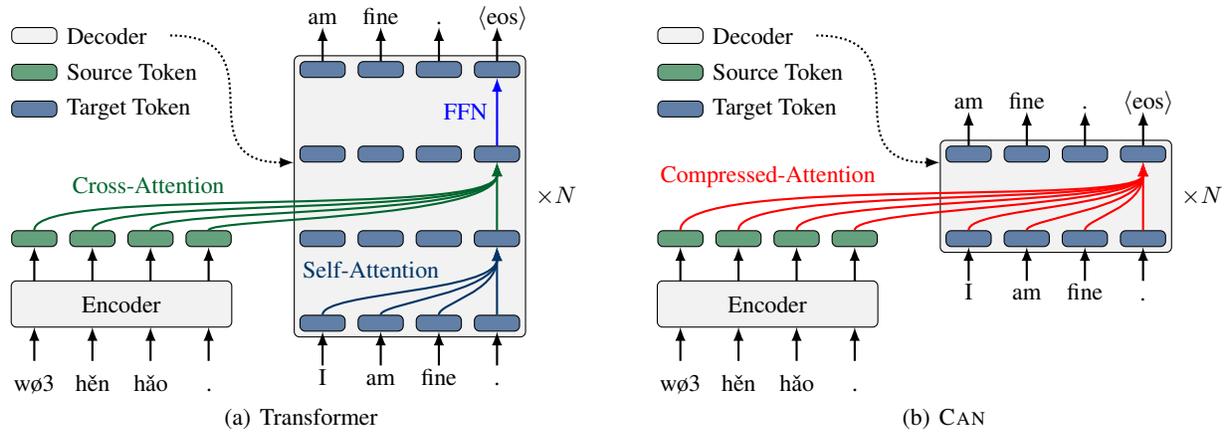

To summarize, our contributions are as follows:
\begin{itemize}
    \item We propose \my{}, a novel architecture that accelerates Transformer by compressing its sub-layers for a higher degree of parallelism. \my{} is easy to be implemented.
    \item Our work is based on a stronger baseline, which is 2$\times$ faster than the widely used standard baseline.
    \item The extensive experiments on 14 WMT machine translation tasks show that \my{} is 1.42$\times$ faster than the stronger baseline and 2.82$\times$ for the standard baseline. \my{} also outperforms other approaches such as \SAN{} and \aan{}. 
\end{itemize}

\section{Background: Transformer}

Transformer is one of the state-of-the-art neural models in machine translation. It consists of a $N$-layer encoder and a $N$-layer decoder, where $N=6$ in most cases. The encoder maps the source sentence to a sequence of continuous representations and the decoder maps these representations to the target sentence. All layers in the encoder or decoder are identical to each other.

The layer in the decoder consists of three sub-layers, including the self-attention, the cross-attention and the feed-forward network (FFN). The self-attention takes the output $X$ of the previous sub-layer as its input and produces a tensor with the same size as its output. It computes the attention distribution $A_x$ and then averages $X$ by $A_x$. We denote the self-attention as $Y_x=\mathrm{Self}(X)$, where $X\in\mathbb{R}^{t\times d}$, $t$ is the target sentence length and $d$ is the dimension of the hidden representation:
\begin{align}
    A_x&=\mathrm{SoftMax}(\frac{XW_{q1}W_{k1}^TX^T}{\sqrt{d}})\label{eqn:self-weight}\\
    Y_x&=A_xXW_{v1}\label{eqn:self-sum}
\end{align}
where $W_{q1},W_{k1},W_{v1}\in\mathbb{R}^{d\times d}$.

The cross-attention is similar to the self-attention, except that it takes the encoder output $H$ as an additional input. We denote the cross-attention as $Y_h=\mathrm{Cross}(X,H)$, where $H\in\mathbb{R}^{s\times d}$, $s$ is the source sentence length:
\begin{align}
    A_h&=\mathrm{SoftMax}(\frac{XW_{q2}W_{k2}^TH^T}{\sqrt{d}})\label{eqn:cross-weight}\\
    Y_h&=A_hHW_{v2}\label{eqn:cross-sum}
\end{align}
where $W_{q2},W_{k2},W_{v2}\in\mathbb{R}^{d\times d}$.

The FFN applies non-linear transformation to its input $X$. We denote the FFN as $Y_f=\mathrm{FFN}(X)$:
\begin{equation}
    Y_f=\mathrm{ReLU}(XW_1+b_1)W_2+b_2\label{eqn:ffn}
\end{equation}
where $W_1\in\mathbb{R}^{d\times 4d}$, $b_1\in\mathbb{R}^{4d}$, $W_2\in\mathbb{R}^{4d\times d}$ and $b_2\in\mathbb{R}^{d}$.

All sub-layers are coupled with the residual connection \cite{DBLP:conf/cvpr/HeZRS16}, i.e., $Y=f(X)+X$ where $f$ could be any sub-layer. Their inputs are also preprocessed by the layer normalization first \cite{DBLP:journals/corr/BaKH16}. \fig{fig:transformer} shows the architecture of Transformer decoder. For more details, we refer the reader to \citet{DBLP:conf/nips/VaswaniSPUJGKP17}.

\section{Compressed Attention Network}

\subsection{Compressing Self-Attention and Cross-Attention}

As suggested by \citet{DBLP:conf/eccv/HuangSLSW16}, the output of one layer in the residual network can be decomposed into the sum of all outputs from previous layers. For the adjacent self-attention and cross-attention, we can write their final output as $Y=X+\mathrm{Self}(X)+\mathrm{Cross}(X',H)$, where $X$ is the input of self-attention and $X'=X+\mathrm{Self}(X)$ is the input of cross-attention. If $X$ and $X'$ are identical, we are able to accelerate the computation of $Y$ by parallelizing these two attentions, as $X'$ do not need to wait $\mathrm{Self}(X)$ to finish.

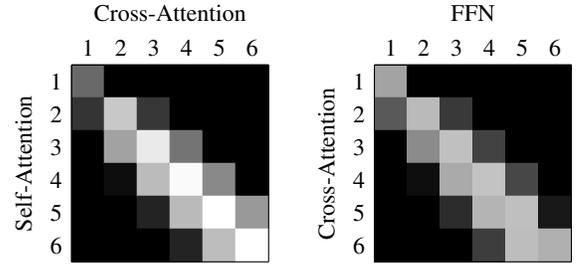
\begin{figure}[t!]
    \centering
    \begin{tikzpicture}
        \begin{groupplot}[
            group style={group size=2 by 1, horizontal sep=40pt},
            width=\linewidth,
            height=0.5\linewidth,
            colormap/blackwhite,
            point meta min=0.7,point meta max=0.95,
        ]
            \nextgroupplot[
                width=0.5\linewidth,height=0.5\linewidth,
                view={0}{90},
                enlargelimits=false,
                ymin=-0.5,ymax=5.5,
                xmin=-0.5,xmax=5.5,
                xlabel={Cross-Attention},
                ylabel={Self-Attention},
                ytick={0,1,2,3,4,5},
                yticklabels={1,2,3,4,5,6},
                xtick={0,1,2,3,4,5},
                xticklabels={1,2,3,4,5,6},
                xtick style={draw=none},
                ytick style={draw=none},
                label style={font=\small},
                every tick label/.append style={font=\small},
                xticklabel pos=upper,
            ]
                \addplot3[matrix plot] table [meta=value] {
                    x y value
                    0 0 0.8025
                    0 1 0.7510
                    0 2 0.5528
                    0 3 0.4033
                    0 4 0.2819
                    0 5 0.1932
                    
                    1 0 0.5666
                    1 1 0.8943
                    1 2 0.8579
                    1 3 0.7125
                    1 4 0.5523
                    1 5 0.4244
                    
                    2 0 0.3827
                    2 1 0.7534
                    2 2 0.9279
                    2 3 0.8817
                    2 4 0.7343
                    2 5 0.5816
                    
                    3 0 0.2676
                    3 1 0.6112
                    3 2 0.8131
                    3 3 0.9429
                    3 4 0.8807
                    3 5 0.7337
                    
                    4 0 0.1797
                    4 1 0.4725
                    4 2 0.6672
                    4 3 0.8337
                    4 4 0.9594
                    4 5 0.8837
                    
                    5 0 0.1242
                    5 1 0.3623
                    5 2 0.5243
                    5 3 0.6894
                    5 4 0.8487
                    5 5 0.9638
                };
            \nextgroupplot[
                width=0.5\linewidth,height=0.5\linewidth,
                view={0}{90},
                enlargelimits=false,
                ymin=-0.5,ymax=5.5,
                xmin=-0.5,xmax=5.5,
                xlabel={FFN},
                ylabel={Cross-Attention},
                ytick={0,1,2,3,4,5},
                yticklabels={1,2,3,4,5,6},
                xtick={0,1,2,3,4,5},
                xticklabels={1,2,3,4,5,6},
                xtick style={draw=none},
                ytick style={draw=none},
                label style={font=\small},
                every tick label/.append style={font=\small},
                xticklabel pos=upper,
            ]
                \addplot3[matrix plot] table [meta=value] {
                    x y value
                    0 0 0.8592
                    0 1 0.7867
                    0 2 0.6084
                    0 3 0.4756
                    0 4 0.3450
                    0 5 0.2484
                    
                    1 0 0.6601
                    1 1 0.8813
                    1 2 0.8360
                    1 3 0.7140
                    1 4 0.5659
                    1 5 0.4297
                    
                    2 0 0.4993
                    2 1 0.7553
                    2 2 0.8870
                    2 3 0.8655
                    2 4 0.7428
                    2 5 0.6011
                    
                    3 0 0.3676
                    3 1 0.6049
                    3 2 0.7645
                    3 3 0.8897
                    3 4 0.8753
                    3 5 0.7593
                    
                    4 0 0.2668
                    4 1 0.4742
                    4 2 0.6232
                    4 3 0.7689
                    4 4 0.8863
                    4 5 0.8839
                    
                    5 0 0.1676
                    5 1 0.3644
                    5 2 0.4760
                    5 3 0.5962
                    5 4 0.7238
                    5 5 0.8716
                };  
        \end{groupplot}
    \end{tikzpicture}
    \caption{The cosine similarity of inputs for every two adjacent sub-layers on WMT14 En-De translation task (a dark cell means the inputs are dissimilar).}
    \label{fig:identity}
\end{figure}

Previous work \cite{DBLP:conf/eccv/HeZRS16} has shown that inputs of adjacent layers are similar. This implies that $X$ and $X'$ are close and the parallelization is possible. We empirically verify this in the left part of \fig{fig:identity} by examining the cosine similarity between inputs of every self-attention and cross-attention pairs. It shows that $X$ and $X'$ are indeed close to each other (a high similarity $>0.9$ for the diagonal entries). Therefore we could assume $X$ and $X'$ are identical (we omit the layer normalization for simplicity):
\begin{equation}
    Y=X+\mathrm{Self}(X)+\mathrm{Cross}(X,H)\label{eqn:self-cross-res}
\end{equation}

By observing that \eqn{eqn:self-sum} and \eqn{eqn:cross-sum} are essentially matrix multiplications, we could rewrite $\mathrm{Self}(X)+\mathrm{Cross}(X,H)$ as a single matrix multiplication:
\begin{align}
    A&=\left[A_x^T,A_h^T\right]^T\\
    \mathrm{Self}(X)+\mathrm{Cross}(X,H)&=A\left[XW_{v1},HW_{v2}\right]\label{eqn:self-cross-sum}
\end{align}
$[\cdot]$ is the concatenation operation along the first dimension.

\citet{DBLP:conf/ijcai/XiaoLZ0L19} shows that some attention distributions $A_x$ and $A_h$ are duplicate. This means that there exists a certain redundancy in $\{W_{q1},W_{k1}\}$ and $\{W_{q2},W_{k2}\}$. Thus we could safely share $W_{q1}$ and $W_{q2}$ to parallelize the computation of the attention distribution $A$:
\begin{align}
    \bar{A}&=\left(XW_q\left[XW_{k1},HW_{k2}\right]^T\right)/\sqrt{d}\\
    A&=\left[\mathrm{SoftMax}(\bar{A}^T_{\cdot,1\ldots t}),\mathrm{SoftMax}(\bar{A}^T_{\cdot,t+1\ldots t+s})\right]^T
\end{align}

However, $A$ consists of two SoftMax distributions and is used in \eqn{eqn:self-cross-sum} without normalization. The output variance is then doubled and leads to poor optimization \cite{DBLP:journals/jmlr/GlorotB10}. It is advised to divide $A$ by $\sqrt{2}$ to preserve the variance. This way resembles a single distribution. So we use one SoftMax instead and this works well:
\begin{equation}
    A=\mathrm{SoftMax}(\frac{XW_q\left[XW_{k1},HW_{k2}\right]^T}{\sqrt{d}})\label{eqn:self-cross-weight}
\end{equation}

Now, we can compute $Y$ in \eqn{eqn:self-cross-res} efficiently by using \eqn{eqn:self-cross-weight} as well as \eqn{eqn:self-cross-sum} to compute $\mathrm{Self}(X)+\mathrm{Cross}(X,H)$.

\begin{figure}[t!]
    \centering
    \begin{tikzpicture}
        \tikzstyle{reprnode} = [rectangle,draw,rounded corners=2pt,minimum height=0.6cm,minimum width=0.2cm,inner sep=3pt]
        \tikzstyle{wordnode} = [minimum width=0.6cm,font=\small,align=center,inner sep=1pt]

        \node[reprnode,fill=ugreen!50!white,anchor=west] (enc1) at (0,0) {};
        \node[reprnode,fill=ugreen!50!white,anchor=west] (enc2) at ([xshift=4pt]enc1.east) {};
        \node[reprnode,fill=ugreen!50!white,anchor=west] (enc3) at ([xshift=4pt]enc2.east) {};
        \node[reprnode,fill=ugreen!50!white,anchor=west] (enc4) at ([xshift=4pt]enc3.east) {};
        \node[reprnode,fill=dblue!50!white,anchor=west] (dec1) at ([xshift=4pt]enc4.east) {};
        \node[reprnode,fill=dblue!50!white,anchor=west] (dec2) at ([xshift=4pt]dec1.east) {};
        \node[reprnode,fill=dblue!50!white,anchor=west] (dec3) at ([xshift=4pt]dec2.east) {};
        \node[reprnode,fill=dblue!50!white,anchor=west] (dec4) at ([xshift=4pt]dec3.east) {};

        \begin{pgfonlayer}{background}
            \node[rectangle,draw,densely dashed,thick,rounded corners=2pt,inner sep=4pt,fit=(enc1) (dec4),label={[font=\small,name=in]below:$\left[H,X\right]$}] (input) {};
        \end{pgfonlayer}
        \draw[decorate,decoration={brace}] (input.south west) to node [midway,left,font=\small] {$d$} (input.north west);

        \begin{pgfonlayer}{background}
            \node[rectangle,draw,densely dashed,thick,rounded corners=2pt,inner sep=2pt,fit=(dec4)] (query) {};
        \end{pgfonlayer}

        \node[reprnode,minimum height=1.2cm,fill=ugreen!50!white,anchor=west] (ev1) at ([xshift=1.5cm]dec4.east) {};
        \node[reprnode,minimum height=1.2cm,fill=ugreen!50!white,anchor=west] (ev2) at ([xshift=4pt]ev1.east) {};
        \node[reprnode,minimum height=1.2cm,fill=ugreen!50!white,anchor=west] (ev3) at ([xshift=4pt]ev2.east) {};
        \node[reprnode,minimum height=1.2cm,fill=ugreen!50!white,anchor=west] (ev4) at ([xshift=4pt]ev3.east) {};
        \node[reprnode,minimum height=1.2cm,fill=dblue!50!white,anchor=west] (dv1) at ([xshift=4pt]ev4.east) {};
        \node[reprnode,minimum height=1.2cm,fill=dblue!50!white,anchor=west] (dv2) at ([xshift=4pt]dv1.east) {};
        \node[reprnode,minimum height=1.2cm,fill=dblue!50!white,anchor=west] (dv3) at ([xshift=4pt]dv2.east) {};
        \node[reprnode,minimum height=1.2cm,fill=dblue!50!white,anchor=west] (dv4) at ([xshift=4pt]dv3.east) {};

        \begin{pgfonlayer}{background}
            \node[rectangle,draw,densely dashed,thick,rounded corners=2pt,inner sep=4pt,fit=(ev1) (dv4),label={[font=\small]below:$Y$}] (v) {};
        \end{pgfonlayer}
        \draw[decorate,decoration={brace,mirror}] (v.south east) to node [midway,right,font=\small] {$4d$} (v.north east);

        \coordinate (p1) at ([yshift=2.2cm]enc1.north);
        \coordinate (p2) at ([yshift=2.2cm+0.1cm]enc2.north);
        \coordinate (p3) at ([yshift=2.2cm+0.3cm]enc3.north);
        \coordinate (p4) at ([yshift=2.2cm+0.7cm]enc4.north);
        \coordinate (p5) at ([yshift=2.2cm+0.7cm]dec1.north);
        \coordinate (p6) at ([yshift=2.2cm+0.3cm]dec2.north);
        \coordinate (p7) at ([yshift=2.2cm+0.1cm]dec3.north);
        \coordinate (p8) at ([yshift=2.2cm]dec4.north);
        \draw[blue,thick,rounded corners=5pt] (p1) -- (p2) -- (p3) -- (p4) -- (p5) -- (p6) -- (p7) -- (p8);
        \draw[->,black,thick] ([yshift=2cm]enc1.north west) to ([yshift=2cm]dec4.north east);
        \node[font=\small,anchor=east] (label) at ([yshift=2.8cm]dec4.north east) {$A$};

        \node[reprnode,minimum width=1.2cm,minimum height=0.2cm,fill=dblue!50!white,anchor=south] (attn1) at ([yshift=1cm]v.north) {};
        \node[reprnode,minimum width=1.2cm,minimum height=0.2cm,fill=dblue!50!white,anchor=south] (attn2) at ([yshift=1cm]attn1.north) {};

        \node[reprnode,minimum width=0.6cm,minimum height=0.2cm,fill=dblue!50!white,anchor=south] (output) at ([yshift=1cm]attn2.north) {};
        \draw[decorate,decoration={brace}] (output.north west) to node [midway,above,font=\small] {$d$} (output.north east);

        \draw[-latex,thick] (input.east) to node [pos=0.45,font=\scriptsize,rotate=-90,fill=white] {$Y=\left[H\widetilde{W}_{v2},X\widetilde{W}_{v1}\right]$} (v.west);

        \draw[-latex,thick] (input.north) to ([yshift=2cm-4pt]input.north);
        \draw[-latex,thick] (query.north) to [out=90,in=-90] ([yshift=2cm-4pt]input.north);
        \node[fill=white,font=\small,anchor=south] () at ([shift={(0.2cm,0.8cm)}]input.north) {\eqn{eqn:self-cross-weight}};

        \draw[-latex,thick] (v.north) to node [midway,pos=0.4,fill=white,font=\small,name=eq1] {$Y=AY$} (attn1.south);

        \draw[-latex,thick] (attn1.north) to node [midway,pos=0.4,fill=white,font=\small,name=eq2] {$Y=XW_1+Y+b_1$} (attn2.south);
        \draw[-latex,thick] (query.north) to [out=90,in=180] (eq2.west);

        \draw[-latex,thick] (label.east) to [out=0,in=180] (eq1.west);

        \draw[-latex,thick] (attn2.north) to node [midway,pos=0.4,fill=white,font=\small] {$Y=\mathrm{ReLU}(Y)W_2+b_2$} (output.south);

        \node[font=\small,anchor=east,text=red] (label1) at ([xshift=-0.5cm]in.west) {Input};
        \draw[->,thick,densely dotted] (label1) to (in);
        \node[font=\small,anchor=east,inner sep=1pt,text=red] (label2) at ([xshift=-0.5cm]output.west) {Output};
        \draw[->,thick,densely dotted] (label2) to (output);
        
        \node[reprnode,minimum height=0.2cm,minimum width=0.6cm,fill=dblue!50!white,anchor=south west,label={[font=\small]right:Target Token $X$}] (label1) at ([yshift=3.7cm]enc1.north west) {};
        \node[reprnode,minimum height=0.2cm,minimum width=0.6cm,fill=ugreen!50!white,anchor=south west,label={[font=\small]right:Source Token $H$}] (label2) at ([yshift=5pt]label1.north west) {};
    \end{tikzpicture}
    \caption{Compressed-Attention.}
    \label{fig:process}
\end{figure}
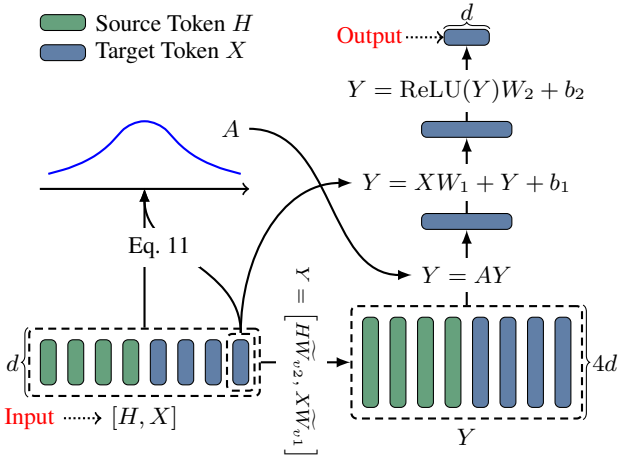

\subsection{Compressing Attention and FFN}

It is natural to consider to merge the attention and FFN with the same approach for further speed-up. As suggested by the right part of \fig{fig:identity}, the similarities between inputs of the adjacent cross-attention and FFN are low (dark diagonal entries). This implies that it is not ideal to make the identical input assumption to parallelize the cross-attention and FFN.

Here we provide another solution. Given that attention is merely a weighted sum and FFN performs a linear projection first, we can merge them by exploiting the linearity. This way not only parallelizes the computation of attention and FFN but also removes redundant matrix multiplications.

We substitute $X$ in \eqn{eqn:ffn} by $Y$ in \eqn{eqn:self-cross-res}:
\begin{equation}
    Y_f=\mathrm{ReLU}(XW_1+A\left[XW_{v1},HW_{v2}\right]W_1+b_1)W_2+b_2
\end{equation}

We can combine $W_1$ with $W_{v1}$ as well as $W_{v2}$ into $\widetilde{W}_{v1},\widetilde{W}_{v2}\in\mathbb{R}^{d\times 4d}$, as these matrices are learnable and matrix multiplied together:
\begin{equation}
    Y_f=\mathrm{ReLU}(XW_1+A\left[X\widetilde{W}_{v1},H\widetilde{W}_{v2}\right]+b_1)W_2+b_2\label{eqn:self-cross-ffn}
\end{equation}
Furthermore, $XW_1$ can be computed in parallel with other transformations such as $XW_q$.

This eventually gives us an more efficient decoder layer architecture, named \emph{Compressed-Attention}. The whole computation process is shown in \fig{fig:process}: it first computes the attention distribution $A$ by \eqn{eqn:self-cross-weight}, then performs the attention operation via \eqn{eqn:self-cross-ffn}, and produces $Y_f$ as the final result. The proposed \emph{Compressed Attention Network} (\my{}) stacks compressed-attentions to form its decoder. \fig{fig:comparison} shows the difference between Transformer and \my{}.

\subsection{Balancing Encoder and Decoder Depths}

Based on the findings of \citet{DBLP:journals/corr/abs-2006-10369}, we learn that a shallow decoder could offer a great speed gain, while a deep encoder could make up of the loss of a shallow decoder without adding a heavy computation overhead. Since their work is based on knowledge distillation \cite{DBLP:journals/corr/HintonVD15}, here we re-examine this idea under the standard training setting (without knowledge distillation).

\begin{figure}[t!]
    \centering
    \begin{tikzpicture}
        \begin{axis}[
            width=0.8\linewidth,height=0.6\linewidth,
            yticklabel style={/pgf/number format/fixed,/pgf/number format/precision=1},
            ymin=26,ymax=28,
            enlarge y limits=0.15,
            ytick={26,27,28},
            ylabel={BLEU [\%]},
            ylabel near ticks,
            xlabel={\# of Encoder Layers/\# of Decoder Layers},
            xlabel near ticks,
            symbolic x coords={6/6,9/4,12/2,14/1},
            xmajorgrids=true,
            ymajorgrids=true,
            grid style=dashed,
            xtick=data,
            every tick label/.append style={font=\small},
            label style={font=\small},
          ]
            \addplot [lyyred,thick,mark=*] coordinates {
                (6/6,27.32) (9/4,27.63) (12/2,27.46) (14/1,26.20)
            };\label{bleu}
        \end{axis}
        \begin{axis}[
            width=0.8\linewidth,height=0.6\linewidth,
            yticklabel style={/pgf/number format/fixed,/pgf/number format/precision=1},
            ymin=100,ymax=300,
            enlarge y limits=0.15,
            xtick=\empty,
            ytick={100,200,300},
            ylabel={Speed},
            ylabel near ticks,
            axis y line*=right,
            symbolic x coords={6/6,9/4,12/2,14/1},
            every tick label/.append style={font=\small},
            label style={font=\small},
            legend columns=2,
            legend pos=north west,
            legend style={font=\small},
          ]
            \addlegendimage{/pgfplots/refstyle=bleu}\addlegendentry{BLEU}
            \addplot [lyyblue,thick,mark=square*] coordinates {
                (6/6,104.27) (9/4,144.16) (12/2,219.53) (14/1,292.49)
            };\addlegendentry{Speed}
        \end{axis}
    \end{tikzpicture}
    \caption{Performance (BLEU) and translation speed (token/sec) vs. the numbers of encoder and decoder layers on WMT14 En-De translation task.}
    \label{fig:balance}
\end{figure}
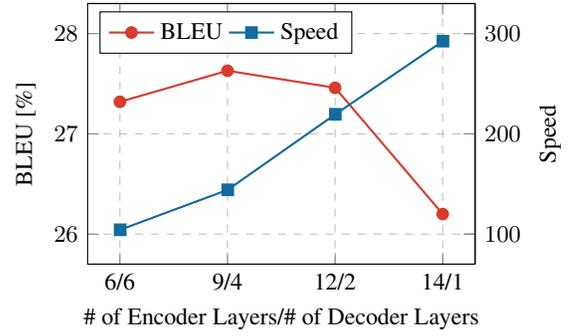

\fig{fig:balance} shows the performance and speed if we gradually reduce the decoder depth while adding more encoder layers. We see that although the overall number of parameters remains the same, the baseline can be 2$\times$ faster without losing any performance (12/2 vs. 6/6). This justifies the previous idea. We thereby choose a stronger baseline with a 12-layer encoder and a 2-layer decoder for a more convincing comparison. This setting is also applied to \my{}.

\section{Experiments}

\subsection{Experimental Setup}

\begin{table}[t!]
    \centering
    \renewcommand\tabcolsep{2.5pt}
    \begin{tabular}{l|l|r|r|r|r|r|r}
        \hline
        \multicolumn{1}{c|}{\multirow{2}*{Source}} & 
        \multicolumn{1}{c|}{\multirow{2}*{Lang.}} & 
        \multicolumn{2}{c|}{Train} & 
        \multicolumn{2}{c|}{Valid} & 
        \multicolumn{2}{c}{Test} \\
        \cline{3-8}
        & &
        \multicolumn{1}{c|}{sent.} & 
        \multicolumn{1}{c|}{word} &
        \multicolumn{1}{c|}{sent.} & 
        \multicolumn{1}{c|}{word} &
        \multicolumn{1}{c|}{sent.} &
        \multicolumn{1}{c}{word} \\
        \hline
        \multirow{2}*{WMT14} & En$\leftrightarrow$De & 4.5M & 220M & 3000 & 110K & 3003 & 114K \\
        \cline{2-8}
        & En$\leftrightarrow$Fr & 35M & 2.2B & 26K & 1.7M & 3003 & 155K \\
        \cline{1-8}
        \multirow{5}*{WMT17} & En$\leftrightarrow$De & 5.9M & 276M & 8171 & 356K & 3004 & 128K \\
        \cline{2-8}
        & En$\leftrightarrow$Fi & 2.6M & 108M & 8870 & 330K & 3002 & 110K \\
        \cline{2-8}
        & En$\leftrightarrow$Lv & 4.5M & 115M & 2003 & 90K & 2001 & 88K \\
        \cline{2-8}
        & En$\leftrightarrow$Ru & 25M & 1.2B & 8819 & 391K & 3001 & 132K \\
        \cline{2-8}
        & En$\leftrightarrow$Cs & 52M & 1.2B & 8658 & 354K & 3005 & 118K \\
        \hline
    \end{tabular}
    \caption{Data statistics (\# of sentences and \# of words).}
    \label{tab:data}
\end{table}

\subsubsection{Datasets}

We evaluate our methods on 14 machine translation tasks (7 datasets $\times$ 2 translation directions each), including WMT14 En$\leftrightarrow$\{De, Fr\} and WMT17 En$\leftrightarrow$\{De, Fi, Lv, Ru, Cs\}.

WMT14 En$\leftrightarrow$\{De, Fr\} datasets are tokenized by a script from Moses\footnote{\url{https://github.com/moses-smt/mosesdecoder/blob/master/scripts/tokenizer/tokenizer.perl}}. We apply BPE \cite{DBLP:conf/acl/SennrichHB16a} with 32K merge operations to segment words into subword units. Sentences with more than 250 subword units are removed. The first two rows of \tab{tab:data} are the detailed statistics of these two datasets. For En-De, we share the source and target vocabularies. We choose \emph{newstest-2013} as the validation set and \emph{newstest-2014} as the test set. For En-Fr, we validate the system on the combination of \emph{newstest-2012} and \emph{newstest-2013}, and test it on \emph{newstest-2014}.

All WMT17 datasets are the official preprocessed version from WMT17 website\footnote{\url{http://data.statmt.org/wmt17/translation-task/preprocessed/}}. BPE with 32K merge operations is similarly applied to these datasets. We use the concatenation of all available preprocessed validation sets in WMT17 datasets as our validation set:
\begin{itemize}
    \item En$\leftrightarrow$De. We use the concatenation of \emph{newstest2014}, \emph{newstest2015} and \emph{newstest2016} as the validation set.
    \item En$\leftrightarrow$Fi. We use the concatenation of \emph{newstest2015}, \emph{newsdev2015}, \emph{newstest2016} and \emph{newstestB2016} as the validation set.
    \item En$\leftrightarrow$Lv. We use \emph{newsdev2016} as the validation set.
    \item En$\leftrightarrow$Ru. We use the concatenation of \emph{newstest2014}, \emph{newstest2015} and \emph{newstest2016} as the validation set.
    \item En$\leftrightarrow$Cs. We use the concatenation of \emph{newstest2014}, \emph{newstest2015} and \emph{newstest2016} as the validation set.
\end{itemize}
We use \emph{newstest2017} as the test set for all WMT17 datasets. Detailed statistics of these datasets are shown in \tab{tab:data}. For all 14 translation tasks, we report case-sensitive tokenized BLEU scores\footnote{\url{https://github.com/moses-smt/mosesdecoder/blob/master/scripts/generic/multi-bleu.perl}}.

\begin{table}[t!]
    \centering
    \renewcommand\tabcolsep{4pt}
    \begin{tabular}{c|l|c|r|c|r|r}
        \hline
        & 
        \multicolumn{1}{c|}{System} &
        \multicolumn{1}{c|}{Test} & 
        \multicolumn{1}{c|}{\small$\Delta_{\mathrm{BLEU}}$} &
        \multicolumn{1}{c|}{Valid} & 
        \multicolumn{1}{c|}{Speed} &
        \multicolumn{1}{c}{\small$\Delta_{\mathrm{Speed}}$} \\
        \hline
        \multirow{5}*{\rotatebox{90}{En-De}} & Baseline & 27.32 & \multicolumn{1}{c|}{-} & 26.56 & 104.27 & \multicolumn{1}{c}{-} \\
        & Balanced & 27.46 & 0.00 & 26.81 & 219.53 & 0.00\% \\
        \cline{2-7}
        & \SAN{} & 26.91 & -0.55 & 26.04 & 229.89 & +4.72\% \\
        & \aan{} & 27.36 & -0.10 & 26.11 & 233.58 & +6.40\% \\
        & \my{} & 27.32 & -0.14 & 26.47 & 290.08 & +32.14\% \\
        \hline
        \multirow{5}*{\rotatebox{90}{De-En}} & Baseline & 30.50 & \multicolumn{1}{c|}{-} & 30.34 & 103.97 & \multicolumn{1}{c}{-} \\
        & Balanced & 30.76 & 0.00 & 30.37 & 206.00 & 0.00\% \\
        \cline{2-7}
        & \SAN{} & 30.09 & -0.67 & 30.11 & 240.52 & +16.76\% \\
        & \aan{} & 30.15 & -0.61 & 30.07 & 232.08 & +12.66\% \\
        & \my{} & 30.37 & -0.39 & 30.17 & 293.16 & +42.31\% \\
        \hline
        \hline
        \multirow{5}*{\rotatebox{90}{En-Fr}} & Baseline & 40.82 & \multicolumn{1}{c|}{-} & 46.80 & 104.65 & \multicolumn{1}{c}{-} \\
        & Balanced & 40.55 & 0.00 & 46.87 & 206.54 & 0.00\% \\
        \cline{2-7}
        & \SAN{} & 40.45 & -0.10 & 46.69 & 208.68 & +1.04\% \\
        & \aan{} & 40.50 & -0.05 & 46.57 & 210.29 & +1.82\% \\
        & \my{} & 40.25 & -0.30 & 46.56 & 263.83 & +27.74\% \\
        \hline
        \multirow{5}*{\rotatebox{90}{Fr-En}} & Baseline & 36.33 & \multicolumn{1}{c|}{-} & 47.03 & 105.85 & \multicolumn{1}{c}{-} \\
        & Balanced & 36.86 & 0.00 & 46.89 & 201.13 & 0.00\% \\
        \cline{2-7}
        & \SAN{} & 36.73 & -0.13 & 46.82 & 213.30 & +6.05\% \\
        & \aan{} & 36.52 & -0.34 & 46.74 & 215.97 & +7.38\% \\
        & \my{} & 36.67 & -0.19 & 46.63 & 266.63 & +32.57\% \\
        \hline
    \end{tabular}
    \caption{Comparison of BLEU scores [\%] and translation speeds (token/sec) of different attention models on WMT14 En$\leftrightarrow$\{De, Fr\} translation tasks.}
    \label{tab:wmt14}
\end{table}

\subsubsection{Model Setup}

Our baseline system is based on the open-source implementation of the Transformer model presented in \citet{DBLP:conf/naacl/OttEBFGNGA19}. For all machine translation tasks, the standard Transformer baseline (Baseline) consists of a 6-layer encoder and a 6-layer decoder. The embedding size is set to 512. The number of attention heads is 8. The FFN hidden size equals to 4$\times$ embedding size. Dropout with the value of 0.1 is used for regularization. We adopt the inverse square root learning rate schedule with 8,000 warmup steps and $0.0007$ learning rate. We stop training until the model stops improving on the validation set. All systems are trained on 8 NVIDIA TITIAN V GPUs with mixed-precision training \cite{DBLP:conf/iclr/MicikeviciusNAD18} and a batch size of 4,096 tokens per GPU. We average model parameters in the last 5 epochs for better performance. At test time, the model is decoded with a beam of width 4 and half-precision. For an accurate speed comparison, we decode with a batch size of 1 to avoid paddings. The stronger balanced baseline (Balanced) shares the setting with this standard baseline, except that its encoder depth is 12 and decoder depth is 2.

We compare \my{} and other model acceleration approaches with our baselines. We choose Sharing Attention Network (\SAN{}) \cite{DBLP:conf/ijcai/XiaoLZ0L19} and Average Attention Network (\aan{}) \cite{DBLP:conf/acl/XiongZS18} for comparison, as they have been proven to be effective in various machine translation tasks \cite{DBLP:conf/aclnmt/BirchFLNO18}. All hyper-parameters of \my{}, \SAN{} and \aan{} are identical to the balanced baseline system. Results are the average of 3 runs.

\begin{table}[t!]
    \centering
    \renewcommand\tabcolsep{4pt}
    \begin{tabular}{c|l|c|r|c|r|r}
        \hline
        & 
        \multicolumn{1}{c|}{System} & 
        \multicolumn{1}{c|}{Test} & 
        \multicolumn{1}{c|}{\small$\Delta_{\mathrm{BLEU}}$} &
        \multicolumn{1}{c|}{Valid} & 
        \multicolumn{1}{c|}{Speed} &
        \multicolumn{1}{c}{\small$\Delta_{\mathrm{Speed}}$} \\
        \hline
        \multirow{3}*{\rotatebox{90}{En-De}} & Baseline & 28.40 & \multicolumn{1}{c|}{-} & 31.30 & 106.58 & \multicolumn{1}{c}{-} \\
        & Balanced & 28.65 & 0.00 & 31.39 & 218.35 & 0.00\% \\
        \cline{2-7}
        & \my{} & 28.30 & -0.35 & 30.94 & 280.57 & +28.50\% \\
        \hline
        \multirow{3}*{\rotatebox{90}{De-En}} & Baseline & 34.48 & \multicolumn{1}{c|}{-} & 35.36 & 103.04 & \multicolumn{1}{c}{-} \\
        & Balanced & 34.38 & 0.00 & 35.16 & 220.05 & 0.00\% \\
        \cline{2-7}
        & \my{} & 33.99 & -0.39 & 34.82 & 286.23 & +30.07\% \\
        \hline
        \hline
        \multirow{3}*{\rotatebox{90}{En-Fi}} & Baseline & 21.28 & \multicolumn{1}{c|}{-} & 18.31 & 103.84 & \multicolumn{1}{c}{-} \\
        & Balanced & 21.38 & 0.00 & 18.67 & 207.73 & 0.00\% \\
        \cline{2-7}
        & \my{} & 21.14 & -0.24 & 18.19 & 286.36 & +37.85\% \\
        \hline
        \multirow{3}*{\rotatebox{90}{Fi-En}} & Baseline & 25.54 & \multicolumn{1}{c|}{-} & 21.32 & 106.59 & \multicolumn{1}{c}{-} \\
        & Balanced & 25.63 & 0.00 & 21.29 & 209.88 & 0.00\% \\
        \cline{2-7}
        & \my{} & 25.25 & -0.38 & 21.31 & 287.57 & +37.02\% \\
        \hline
        \hline
        \multirow{3}*{\rotatebox{90}{En-Lv}} & Baseline & 16.14 & \multicolumn{1}{c|}{-} & 21.33 & 107.20 & \multicolumn{1}{c}{-} \\
        & Balanced & 15.98 & 0.00 & 21.21 & 219.02 & 0.00\% \\
        \cline{2-7}
        & \my{} & 15.90 & -0.08 & 20.75 & 287.33 & +31.19\% \\
        \hline
        \multirow{3}*{\rotatebox{90}{Lv-En}} & Baseline & 18.74 & \multicolumn{1}{c|}{-} & 24.79 & 106.25 & \multicolumn{1}{c}{-} \\
        & Balanced & 18.69 & 0.00 & 24.54 & 216.06 & 0.00\% \\
        \cline{2-7}
        & \my{} & 18.21 & -0.48 & 24.16 & 275.89 & +27.69\% \\
        \hline
        \hline
        \multirow{3}*{\rotatebox{90}{En-Ru}} & Baseline & 30.44 & \multicolumn{1}{c|}{-} & 30.67 & 106.46 & \multicolumn{1}{c}{-} \\
        & Balanced & 30.28 & 0.00 & 30.59 & 214.52 & 0.00\% \\
        \cline{2-7}
        & \my{} & 29.89 & -0.39 & 30.28 & 287.13 & +33.85\% \\
        \hline
        \multirow{3}*{\rotatebox{90}{Ru-En}} & Baseline & 34.44 & \multicolumn{1}{c|}{-} & 32.39 & 107.24 & \multicolumn{1}{c}{-} \\
        & Balanced & 34.24 & 0.00 & 32.22 & 213.78 & 0.00\% \\
        \cline{2-7}
        & \my{} & 33.95 & -0.29 & 31.92 & 287.86 & +34.65\% \\
        \hline
        \hline
        \multirow{3}*{\rotatebox{90}{En-Cs}} & Baseline & 24.00 & \multicolumn{1}{c|}{-} & 28.09 & 106.18 & \multicolumn{1}{c}{-} \\
        & Balanced & 23.69 & 0.00 & 28.03 & 212.65 & 0.00\% \\
        \cline{2-7}
        & \my{} & 23.59 & -0.10 & 27.71 & 272.37 & +28.08\% \\
        \hline
        \multirow{3}*{\rotatebox{90}{Cs-En}} & Baseline & 30.00 & \multicolumn{1}{c|}{-} & 33.01 & 104.00 & \multicolumn{1}{c}{-} \\
        & Balanced & 30.06 & 0.00 & 32.86 & 202.96 & 0.00\% \\
        \cline{2-7}
        & \my{} & 29.87 & -0.19 & 32.99 & 269.70 & +32.88\% \\
        \hline
    \end{tabular}
    \caption{BLEU scores [\%] and translation speeds (token/sec) on WMT17 En$\leftrightarrow$\{De, Fi, Lv, Ru, Cs\} translation tasks.}
    \label{tab:wmt17}
\end{table}

\subsection{Results}

\tab{tab:wmt14} shows the results of various systems on WMT14 En$\leftrightarrow$\{De, Fr\}. Our balanced baseline has nearly the same performance as the standard baseline, but its speed is 2$\times$ faster on average. A similar phenomenon is also observed from WMT17 experiments in \tab{tab:wmt17}. This observation indicates that existing systems do not well balance the encoder and decoder depths. We also report the performance of \aan{}, \SAN{} and the proposed \my{}. All three approaches have similar BLEU scores and slightly underperform the balanced baseline. \my{} is more stable than the others, as its maximum $\Delta_{\mathrm{BLEU}}$ is -0.39, while \SAN{} is -0.67 and \aan{} is -0.61. For speeds of these systems, \SAN{} and \aan{} have a similar level of acceleration (1$\sim$16\%) over the balanced baseline. \my{}, on the other hand, provides a higher level of acceleration (27$\sim$42\%). Interestingly, we find that the acceleration is more obvious in De-En than in others, e.g., 42\% in De-En and 27\% in En-Fr for \my{}. We find that the length ratio between the translation and the source sentence in De-En is higher than others, e.g., 1.0 for De-En and 0.981 for En-Fr. In this case the decoder tends to predict more words and consumes more time in De-En, and thus acceleration approaches that work on the decoder are more effective. In addition, though not reported in \tab{tab:wmt14}, we find that applying \my{} on the standard baseline hurt the performance less (-0.09 BLEU points on average) than on the balanced baseline (-0.25 BLEU points on average).

More experimental results to justify the effectiveness of \my{} are presented in \tab{tab:wmt17}. We evaluate the balanced baseline as well as \my{} on five WMT17 language pairs. The results again show that the balanced baseline is indeed a strong baseline with BLEU scores close to the standard baseline and is consistently 2$\times$ faster. \my{} also shows a similar trend that it slightly underperforms the balanced baseline ($<0.5$ BLEU scores) but is $>27\%$ faster.

\section{Analysis}

\begin{figure*}[t!]
    \centering
    \begin{tikzpicture}
        \begin{groupplot}[
            group style={group size=3 by 1, horizontal sep=50pt},
            width=\linewidth,
            height=0.25\linewidth,
            legend style={
                column sep=10pt,
                legend columns=1,
            },
            legend cell align={left},
        ]
            \nextgroupplot[
                width=0.32\linewidth,height=0.25\linewidth,
                yticklabel style={/pgf/number format/fixed,/pgf/number format/precision=1},
                ymax=450,
                ylabel={Speed},
                ylabel near ticks,
                xlabel={Beam Size},
                xlabel near ticks,
                enlargelimits=0.05,
                symbolic x coords={1,4,16,64,256},
                xmajorgrids=true,
                ymajorgrids=true,
                grid style=dashed,
                xtick=data,
                every tick label/.append style={font=\small},
                label style={font=\small},
                ylabel style={yshift=5pt},
                legend pos=north east,
                legend style={column sep=3pt},
            ]
                \addplot [lyyred,thick,mark=square*] coordinates {
                    (1,259.02) (4,219.53) (16,145.62) (64,46.50) (256,12.43)
                };\addlegendentry{Balanced}
                \addplot [lyyblue,thick,thick,mark=triangle*] coordinates {
                    (1,359.40) (4,290.08) (16,168.54) (64,48.63) (256,15.28)
                };\addlegendentry{\my{}}
            \nextgroupplot[
                width=0.32\linewidth,height=0.25\linewidth,
                yticklabel style={/pgf/number format/fixed,/pgf/number format/precision=1},
                ylabel={Speed},
                ylabel near ticks,
                xlabel={Length},
                xlabel near ticks,
                enlargelimits=0.05,
                symbolic x coords={10,20,30,40,50+},
                xmajorgrids=true,
                ymajorgrids=true,
                grid style=dashed,
                xtick=data,
                every tick label/.append style={font=\small},
                label style={font=\small},
                ylabel style={yshift=5pt},
                legend pos=south east,
                legend style={column sep=3pt},
            ]
                \addplot [lyyred,thick,mark=square*] coordinates {
                    (10,51.33) (20,126.07) (30,134.39) (40,176.06) (50+,186.26)
                };\addlegendentry{Balanced}
                \addplot [lyyblue,thick,thick,mark=triangle*] coordinates {
                    (10,69.71) (20,161.19) (30,201.81) (40,218.78) (50+,237.85)
                };\addlegendentry{\my{}}
            \nextgroupplot[
                width=0.32\linewidth,height=0.25\linewidth,
                symbolic x coords={CAN,SAN,AAN},
                ymin=26,
                enlarge x limits=0.3,
                enlarge y limits={upper,value=1.1},
                ylabel={Translation Length},
                ylabel near ticks,
                ybar=0pt,
                xtick=data,
                ytick=\empty,
                nodes near coords,
                every node near coord/.append style={rotate=90,anchor=west,font=\small,/pgf/number format/fixed,/pgf/number format/fixed zerofill,/pgf/number format/precision=1},
                every tick label/.append style={font=\small},
                xticklabel style={rotate=45,anchor=north east,font=\small,inner sep=0pt,outer sep=2pt},
                bar width=10pt,
                xmajorgrids=true,
                ymajorgrids=true,
                grid style=dashed,
                label style={font=\small},
                legend pos=north east,
                legend image code/.code={
                    \draw [#1] (0cm,-0.15cm) rectangle (0.4cm,0.15cm);
                },
                legend columns=2,
                legend style={at={(0.5,0.98)},anchor=north,column sep=3pt},
            ]
                \addplot [draw=lyyblue!80,fill=lyyblue!60,pattern=north west lines,pattern color=lyyblue] coordinates {
                    (CAN,26.94) (SAN,26.97) (AAN,27.07)
                };\addlegendentry{En-De}
                \addplot [draw=lyygreen!80,fill=lyygreen!60,pattern=north east lines,pattern color=lyygreen] coordinates {
                    (CAN,30.32) (SAN,30.43) (AAN,30.43)
                };\addlegendentry{En-Fr}
        \end{groupplot}
    \end{tikzpicture}
    \caption{Translation speed (token/sec) vs. beam size and translation length on WMT14 En-De translation task.}
    \label{fig:sensitivity}
\end{figure*}
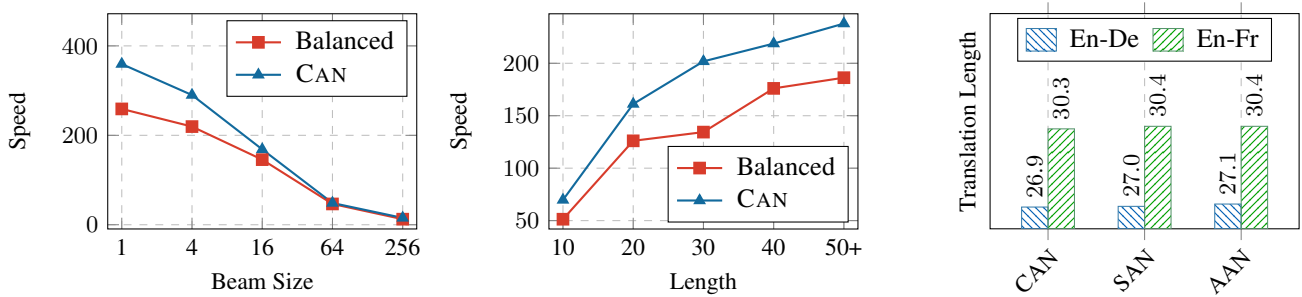

\subsection{Knowledge Distillation}

\begin{table}[t!]
    \centering
    \begin{tabular}{l|c|r|c|r}
            \hline
            \multicolumn{1}{c|}{\multirow{2}*{System}} & 
            \multicolumn{2}{c|}{Before KD} & 
            \multicolumn{2}{c}{After KD} \\
            \cline{2-5}
            &
            \multicolumn{1}{c|}{Test} & 
            \multicolumn{1}{c|}{\small$\Delta_{\mathrm{BLEU}}$} &
            \multicolumn{1}{c|}{Test} &
            \multicolumn{1}{c}{\small$\Delta_{\mathrm{BLEU}}$} \\
            \hline
            Balanced & 27.46 & 0.00 & 27.82 & 0.00 \\
            \hline
            \SAN{} & 26.91 & -0.55 & 27.76 & -0.06 \\
            \aan{} & 27.36 & -0.10 & 27.85 & +0.03 \\
            \my{} & 27.32 & -0.14 & 28.08 & +0.26 \\
            \hline
    \end{tabular}
    \caption{BLEU scores [\%] of applying knowledge distillation (KD) on WMT14 En-De translation task.}
    \label{tab:kd}
\end{table}

Although \SAN{}, \aan{} and \my{} offer considerable speed gain over the balanced baseline, they all suffer from the performance degradation as shown in \tab{tab:wmt14} and \tab{tab:wmt17}. The popular solution to this is knowledge distillation (KD). Here we choose sequence-level knowledge distillation \cite{DBLP:conf/emnlp/KimR16} for better performance in machine translation tasks. The balanced baseline is used to generate the pseudo data for KD.

\tab{tab:kd} shows that KD closes the performance gap between the fast attention models (\SAN{}, \aan{} and \my{}) and the balanced baseline. This fact suggests that all three systems have enough capacity for a good performance, but training from scratch is not able to reach a good convergence state. It suggests that these systems might require a more careful hyper-parameters tuning or a better optimization method.

\subsection{Ablation Study}

\begin{table}[t!]
    \centering
    \renewcommand\tabcolsep{3pt}
    \begin{tabular}{l|c|r|r|r}
        \hline
        \multicolumn{1}{c|}{System} & 
        \multicolumn{1}{c|}{Test} & 
        \multicolumn{1}{c|}{\small$\Delta_{\mathrm{BLEU}}$} &
        \multicolumn{1}{c|}{Speed} &
        \multicolumn{1}{c}{\small$\Delta_{\mathrm{Speed}}$} \\
        \hline
        Balanced & 27.46 & 0.00 & 219.53 & 0.00\% \\
        \hline
        + Compress Attention & 27.09 & -0.37 & 263.64 & +20.09\% \\
        + Compress FFN & 27.69 & +0.23 & 233.17 & +6.21\% \\
        + Compress All & 27.32 & -0.14 & 290.08 & +32.14\% \\
        \hline
    \end{tabular}
    \caption{Ablation study on WMT14 En-De translation task (Compress Attention: compress the self-attention and cross-attention only; Compress FFN: compress the cross-attention and FFN only; Compress All: compress the self-attention, cross-attention and FFN).}
    \label{tab:ablation}
\end{table}

To investigate in which part \my{} contributes the most to the acceleration as well as the performance loss, we only compress the self-attention and cross-attention or compress the cross-attention and FFN for study. \tab{tab:ablation} shows the results of this ablation study. We can see that compressing the two attentions provides a 20.09\% speed-up, while only 6.21\% for compressing attention and FFN. This is because FFN is already highly parallelized and accelerating itself does not bring much gain. Note that the second row of \tab{tab:ablation} is exactly \citet{DBLP:conf/emnlp/ZhangTS19}'s work (without AAN). We see that \my{} is faster and performs better. On the other hand, compressing attentions brings the most performance loss, which shows that the identical input assumption is strong. \fig{fig:identity} shows that inputs of the adjacent layers are not very similar in lower layers. Therefore using \my{} in low layers might bring a great loss. We also find that compressing attention and FFN has an even better result. This might be that we remove the redundant parameters in the model.

\subsection{Sensitivity Analysis}

We study how the speed could be affected by other factors in \fig{fig:sensitivity}, e.g., the beam size and the translation length. The left of \fig{fig:sensitivity} shows that \my{} is consistently faster than the balanced baseline with different beam size. As the acceleration provided by \my{} is constantly proportional to the speed of the baseline, it becomes less obvious when the baseline is slow, i.e., translating with a large beam. An opposite trend happens in the middle of \fig{fig:sensitivity} for the translation length. This is because overheads such as data preparation dominate the translation time of short sentences. This way results in a slow speed even when the translation time is short. As both the baseline and \my{} get faster when generating longer translations, one might suspect that the superior acceleration of \my{} over other approaches comes from the fact that \my{} generates longer translations. Further analysis is conducted and shown in the right of \fig{fig:sensitivity}. We see that \my{}, \SAN{} and \aan{} generate translations with similar lengths in two WMT14 translation tasks. This observation justifies that the superior acceleration brought by \my{} did come from its design rather than translation lengths.

\subsection{Error Analysis}

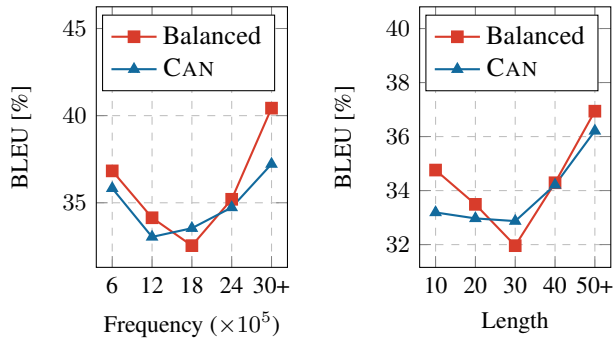
\begin{figure}
    \centering
    \begin{tikzpicture}
        \begin{groupplot}[
            group style={group size=2 by 1, horizontal sep=50pt},
            width=\linewidth,
            height=0.7\linewidth,
        ]
            \nextgroupplot[
                width=0.49\linewidth,height=0.6\linewidth,
                yticklabel style={/pgf/number format/fixed,/pgf/number format/fixed zerofill,/pgf/number format/precision=0},
                ylabel={BLEU [\%]},
                ylabel near ticks,
                xlabel={Frequency ($\times 10^5$)},
                xlabel near ticks,
                enlargelimits=0.1,
                ymax=45,
                symbolic x coords={6,12,18,24,30+},
                xmajorgrids=true,
                ymajorgrids=true,
                grid style=dashed,
                xtick=data,
                every tick label/.append style={font=\small},
                label style={font=\small},
                ylabel style={yshift=5pt},
                legend style={legend columns=1},
                legend pos=north west,
                legend cell align={left},
            ]
                \addplot [lyyred,thick,mark=square*] coordinates {
                    (6,36.83) (12,34.14) (18,32.54) (24,35.20) (30+,40.43)
                };\addlegendentry{Balanced}
                \addplot [lyyblue,thick,thick,mark=triangle*] coordinates {
                    (6,35.83) (12,33.04) (18,33.54) (24,34.73) (30+,37.22)
                };\addlegendentry{\my{}}
            \nextgroupplot[
                width=0.49\linewidth,height=0.6\linewidth,
                yticklabel style={/pgf/number format/fixed,/pgf/number format/fixed zerofill,/pgf/number format/precision=0},
                ylabel={BLEU [\%]},
                ylabel near ticks,
                xlabel={Length},
                xlabel near ticks,
                enlargelimits=0.1,
                ymax=40,
                symbolic x coords={10,20,30,40,50+},
                xmajorgrids=true,
                ymajorgrids=true,
                grid style=dashed,
                xtick=data,
                every tick label/.append style={font=\small},
                label style={font=\small},
                ylabel style={yshift=5pt},
                legend style={legend columns=1},
                legend pos=north west,
                legend cell align={left},
            ]
                \addplot [lyyred,thick,mark=square*] coordinates {
                    (10,34.76) (20,33.49) (30,31.96) (40,34.29) (50+,36.94)
                };\addlegendentry{Balanced}
                \addplot [lyyblue,thick,thick,mark=triangle*] coordinates {
                    (10,33.19) (20,32.97) (30,32.87) (40,34.21) (50+,36.21)
                };\addlegendentry{\my{}}           
        \end{groupplot}
    \end{tikzpicture}
    \caption{BLEU score [\%] vs. word frequency ($\times 10^5$) and translation length on WMT14 En-De translation task.}
    \label{fig:error}
\end{figure}

As shown in \tab{tab:wmt14} and \tab{tab:wmt17}, the acceleration of \my{} comes at the cost of performance. Here we conduct experiments to better understand in which aspect \my{} scarifies for speed-up. We first evaluate the sentence-level BLEU score for each translation, then cluster these translations according to their averaged word frequencies or lengths.

\fig{fig:error} shows the results. The left of \fig{fig:error} indicates that \my{} did well on sentences with low frequencies, but not on those with high frequencies. The right of \fig{fig:error} shows that \my{} does not translate short sentences well but is quite good at translating long sentences. These facts are counterintuitive as one might expect a poor model could do well on easy samples (high frequency and short sentences) but not on hard ones (low frequency and long sentences). This might due to the identical input assumptions we used to derive \my{} are critical to easy samples. We left this for the future exploration.

\subsection{Parallelism Study}

\begin{figure}[t!]
    \centering
    \begin{tikzpicture}
        \begin{axis}[
            width=0.8\linewidth,height=0.6\linewidth,
            yticklabel style={/pgf/number format/fixed,/pgf/number format/precision=1},
            scaled y ticks=false,
            enlarge y limits=0.15,
            ylabel={Speed},
            ylabel near ticks,
            xlabel={Batch Size},
            xlabel near ticks,
            xmajorgrids=true,
            ymajorgrids=true,
            grid style=dashed,
            xmin=-10,xmax=70,
            xtick={0,20,40,60},
            ymin=0,ymax=10000,
            ytick={0,2000,4000,6000,8000,10000},
            every tick label/.append style={font=\small},
            label style={font=\small},
          ]
            \addplot [lyyred,thick,mark=square*] coordinates {
                (1,219.53) (2,401.33) (4,778.29) (8,1422.26) (16,2570.12) (32,4491.53) (64,8033.22) 
            };\label{baseline}
            \addplot [lyyblue,thick,mark=triangle*] coordinates {
                (1,290.08) (2,529.92)(4,992.64) (8,1818.88) (16,3062.25) (32,5586.18) (64,8232.19) 
            };\label{ours}
        \end{axis}
        \begin{axis}[
            width=0.8\linewidth,height=0.6\linewidth,
            yticklabel style={/pgf/number format/fixed,/pgf/number format/precision=1},
            enlarge y limits=0.15,
            xtick=\empty,
            xmin=-10,xmax=70,
            ylabel={$\Delta_\mathrm{Speed}$ [\%]},
            ylabel near ticks,
            ymin=0,ymax=50,
            ytick={0,10,20,30,40,50},
            axis y line*=right,
            every tick label/.append style={font=\small},
            label style={font=\small},
            legend columns=2,
            legend pos=north west,
            legend style={font=\small},
            legend cell align={left},
          ]
            \addlegendimage{/pgfplots/refstyle=baseline}\addlegendentry{Balanced}
            \addlegendimage{/pgfplots/refstyle=ours}\addlegendentry{\my{}}
            \addplot [lyygreen,thick,densely dashed,mark=*] coordinates {
                (1,32.14) (2,32.04) (4,27.54) (8,27.89) (16,19.15) (32,24.37) (64,2.48) 
            };\addlegendentry{$\Delta_{\mathrm{Speed}}$}
        \end{axis}
    \end{tikzpicture}
    \caption{Speed (token/sec) and $\Delta_{\mathrm{Speed}}$ [\%] vs. batch size on WMT14 En-De translation task.}
    \label{fig:parallel}
\end{figure}
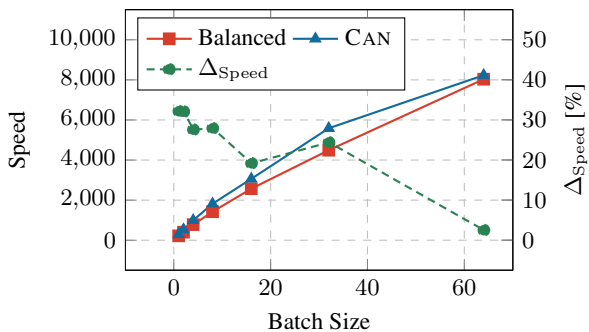

A simple approach to obtain a higher parallelism without modifying the architecture is to increase the batch size at inference. \fig{fig:parallel} compares the inference time of the balanced baseline and \my{} by varying the batch size. We can see that both systems run faster with a larger batch size and \my{} is consistently faster than the balanced baseline. But the acceleration of \my{} over the baseline $\Delta_{\mathrm{Speed}}$ diminishes when the batch size gets larger. In this case we observe that \my{} reaches the highest parallelism (a nearly 100\% GPU utility) in a smaller batch size ($\geq 32$) than the baseline ($>64$). This means that enlarging the batch size no longer provides acceleration for \my{}, while the baseline can still be further speeded up. We expect \my{} could be faster if more tensor cores are available in the future.

\subsection{Training Study}

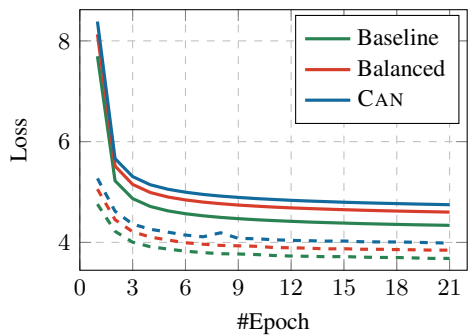
\begin{figure}
    \centering
    \begin{tikzpicture}
        \begin{axis}[
            width=0.8\linewidth,height=0.6\linewidth,
            yticklabel style={/pgf/number format/fixed,/pgf/number format/precision=1},
            ylabel={Loss},
            ylabel near ticks,
            xlabel={\#Epoch},
            xlabel near ticks,
            enlargelimits=0.05,
            xtick distance=3,
            xmajorgrids=true,
            ymajorgrids=true,
            grid style=dashed,
            every tick label/.append style={font=\small},
            label style={font=\small},
            ylabel style={yshift=5pt},
            legend style={font=\small,inner sep=3pt},
            legend image post style={scale=1},
            legend columns=1,
            legend cell align={left},
        ]
            \addplot [lyygreen,very thick] coordinates {
                (1,7.695) (2,5.222) (3,4.870) (4,4.713) (5,4.626) (6,4.569) (7,4.528) (8,4.496) (9,4.471) (10,4.450) (11,4.433) (12,4.418) (13,4.404) (14,4.393) (15,4.383) (16,4.373) (17,4.365) (18,4.357) (19,4.350) (20,4.344) (21,4.338)
            };
            \addlegendentry{Baseline}
            \addplot [lyyred,very thick] coordinates {
                (1,8.129) (2,5.515) (3,5.151) (4,4.990) (5,4.901) (6,4.843) (7,4.800) (8,4.768) (9,4.741) (10,4.720) (11,4.701) (12,4.685) (13,4.672) (14,4.660) (15,4.649) (16,4.639) (17,4.630) (18,4.622) (19,4.615) (20,4.608) (21,4.601)
            };
            \addlegendentry{Balanced}
            \addplot [lyyblue,very thick] coordinates {
                (1,8.384) (2,5.663) (3,5.306) (4,5.144) (5,5.055) (6,4.996) (7,4.952) (8,4.919) (9,4.893) (10,4.870) (11,4.852) (12,4.835) (13,4.821) (14,4.809) (15,4.797) (16,4.787) (17,4.778) (18,4.770) (19,4.762) (20,4.755) (21,4.748)
            };
            \addlegendentry{\my{}}
            \addplot [lyygreen,dashed,very thick] coordinates {
                (1,4.759) (2,4.220) (3,4.004) (4,3.914) (5,3.868) (6,3.825) (7,3.797) (8,3.775) (9,3.770) (10,3.759) (11,3.740) (12,3.729) (13,3.724) (14,3.715) (15,3.719) (16,3.707) (17,3.701) (18,3.700) (19,3.691) (20,3.682) (21,3.682)
            };
            \addplot [lyyred,dashed,very thick] coordinates {
                (1,5.057) (2,4.451) (3,4.211) (4,4.106) (5,4.047) (6,3.993) (7,3.964) (8,3.940) (9,3.932) (10,3.923) (11,3.898) (12,3.893) (13,3.885) (14,3.871) (15,3.874) (16,3.864) (17,3.863) (18,3.857) (19,3.853) (20,3.846) (21,3.846)
            };
            \addplot [lyyblue,dashed,very thick] coordinates {
                (1,5.270) (2,4.619) (3,4.362) (4,4.261) (5,4.203) (6,4.145) (7,4.111) (8,4.191) (9,4.080) (10,4.072) (11,4.053) (12,4.040) (13,4.033) (14,4.025) (15,4.027) (16,4.013) (17,4.009) (18,4.008) (19,4.003) (20,3.993) (21,3.986)
            };
        \end{axis}
    \end{tikzpicture}
    \caption{Loss vs. \# of epochs on WMT14 En-De translation task (solid lines are the training losses, dashed lines are the validation losses).}
    \label{fig:curve}
\end{figure}

We plot the training and validation loss curve of the standard baseline, the balanced baseline and \my{} in \fig{fig:curve} for studying their convergence. We can see that all systems converge stably. The balanced baseline has a higher loss than the standard baseline in both the training and validation sets, but their BLEU scores are close as shown in \tab{tab:wmt14}. This is due to the shallow decoder in the balanced baseline. Since the loss is determined by the decoder, a shallow decoder with less capacity would have a higher loss. \citet{DBLP:conf/acl/WangLXZLWC19} indicates that the encoder depth has a greater impact than the decoder on BLEU scores, therefore the deep encoder makes up the performance loss of the shallow decoder. We also see that \my{} has a higher loss than the balanced baseline because we compress the decoder. Since we do not enhance the encoder, the BLEU score drops accordingly.

\section{Related Work}

\subsection{Model Acceleration}

Large Transformer has demonstrated its effectiveness on various natural language processing tasks, including machine translation \cite{DBLP:conf/nips/VaswaniSPUJGKP17}, language modelling \cite{DBLP:conf/iclr/BaevskiA19} and etc. The by-product brought by this huge network is the slow inference speed. Previous work focuses on improving model efficiency from different perspectives. For example, knowledge distillation approaches treat the large network output as the ground truth to train a small network \cite{DBLP:conf/emnlp/KimR16}. Low-bit quantization approaches represent and run the model with 8-bit integer \cite{DBLP:conf/ijcai/LinLLXLZ20}. Our work follows another line of researches, which purses a more efficient architecture.

\citet{DBLP:conf/acl/FosterVUMKFJSWB18} show that the attention of Transformer benefits the encoder the most and the decoder could be safely replaced by a recurrent network. This way reduces the complexity of the decoder to linear time but incurs a high cost in training. \citet{DBLP:conf/acl/XiongZS18} show that the self-attention is not necessary and a simple averaging is enough. \citet{DBLP:conf/ijcai/XiaoLZ0L19} indicate that most attention distributions are redundant and thus share these distributions among layers. \citet{DBLP:conf/iclr/KitaevKL20} use locality-sensitive hashing to select a constant number of words and perform attention on them. \citet{DBLP:conf/iclr/FanGJ20} train a large Transformer and drop some layers at testing for fast inference. \citet{DBLP:conf/iclr/Gu0XLS18} use a non-autoregressive decoder to predict the whole sentence at one time instead of generating it word by word. This approach makes a linear time translation process to constant time via the parallel computation. Perhaps the most related works are \citet{DBLP:conf/nips/HeTXHQ0L18,DBLP:conf/emnlp/ZhangTS19}. They merge the self-attention and cross-attention and share their parameters. We, on the other hand, use different sets of parameters for each attention and mathematically prove that this way is equivalent to the standard Transformer under some mild conditions. We further show that the attention and FFN could also be merged together due to their linearity.

\subsection{Deep Transformer}

Recent studies have shown that deepening the Transformer encoder is more beneficial than widening the encoder or deepening the decoder \cite{DBLP:conf/emnlp/BapnaCFCW18}. \citet{DBLP:conf/acl/WangLXZLWC19} show that placing the layer normalization before (Pre-Norm) rather than behind (Post-Norm) the sub-layer allows us to train deep Transformer. \citet{DBLP:journals/corr/abs-2002-04745} prove that the success of the Pre-Norm network relies on its well-behaved gradient. \citet{DBLP:conf/emnlp/ZhangTS19} suggest that a proper initialization is enough to train a deep Post-Norm network. \citet{DBLP:journals/corr/abs-2006-10369} similarly exploit this observation but to build a faster instead of a better model. They show that using knowledge distillation, a deep encoder and shallow decoder model could run much faster without losing any performance. Based on their work, we use this model as our baseline system and evaluate it on extensive machine translation tasks without knowledge distillation.

\section{Conclusion}

In this work, we propose \my{}, whose decoder layer consists of only one attention. \my{} offers consistent acceleration by providing a high degree of parallelism. Experiments on 14 WMT machine translation tasks show that \my{} is 2.82$\times$ faster than the baseline. We also use a stronger baseline for comparison. It employs a deep encoder and a shallow decoder, and is 2$\times$ faster than the standard Transformer baseline without loss in performance.

\section*{Acknowledgments}

This work was supported in part by the National Science Foundation of China (Nos. 61876035 and 61732005), and the National Key R\&D Program of China (No. 2019QY1801). The authors would like to thank anonymous reviewers for their comments.

\bibliography{aaai21}

\end{document}